\documentclass{article}

\usepackage{PRIMEarxiv}

\usepackage[utf8]{inputenc} 
\usepackage[T1]{fontenc}    
\usepackage{hyperref}       
\usepackage{url}            
\usepackage{booktabs}       
\usepackage{amsfonts}       
\usepackage{nicefrac}       
\usepackage{microtype}      
\usepackage{lipsum}
\usepackage{fancyhdr}       
\usepackage{graphicx}       
\usepackage{authblk} 
\graphicspath{{media/}}     

\usepackage{caption}
\title{Probing Multimodal Fusion in the Brain: The Dominance of Audiovisual Streams in Naturalistic Encoding
}

\author[1,a]{Hamid Abdollahi}
\author[1,b]{Amir Hossein Mansouri Majoumerd}
\author[1,c]{Amir Hossein Bagheri Baboukani}
\author[1,*]{Amir Abolfazl Suratgar}
\author[1,d]{Mohammad Bagher Menhaj}

\affil[1]{Distributed and Intelligent Optimization Research Laboratory, Electrical Engineering Department, Amirkabir University of Technology, Tehran, Iran}

\affil[ ]{%
	\small\texttt{\{%
		\textsuperscript{a}hamid.abdollahi, 
		\textsuperscript{b}a.mansouri, 
		\textsuperscript{c}amir.b, 
		\textsuperscript{d}menhaj%
		\}@aut.ac.ir}%
}
\affil[*]{\small\texttt{Corresponding author: a-suratgar@aut.ac.ir}}

\begin{document}
\maketitle

\begin{abstract}
	Predicting brain activity in response to naturalistic, multimodal stimuli is a key challenge in computational neuroscience. While encoding models are becoming more powerful, their ability to generalize to truly novel contexts remains a critical, often untested, question. In this work, we developed brain encoding models using state-of-the-art visual (X-CLIP) and auditory (Whisper) feature extractors and rigorously evaluated them on both in-distribution (ID) and diverse out-of-distribution (OOD) data. Our results reveal a fundamental trade-off between model complexity and generalization: a higher-capacity attention-based model excelled on ID data, but a simpler linear model was more robust, outperforming a competitive baseline by 18\% on the OOD set. Intriguingly, we found that linguistic features did not improve predictive accuracy, suggesting that for familiar languages, neural encoding may be dominated by the continuous visual and auditory streams over redundant textual information. Spatially, our approach showed marked performance gains in the auditory cortex, underscoring the benefit of high-fidelity speech representations. Collectively, our findings demonstrate that rigorous OOD testing is essential for building robust neuro-AI models and provides nuanced insights into how model architecture, stimulus characteristics, and sensory hierarchies shape the neural encoding of our rich, multimodal world.
\end{abstract}

\keywords{Brain Encoding Models \and Multimodal Perception \and fMRI \and Naturalistic Stimuli \and Algonauts}

\section{Introduction}
\label{sec:introduction}

The human brain seamlessly integrates sight, sound, and language into a coherent experience of the world. However, how these distinct modalities are represented and combined across the cortex during rich, naturalistic perception remains poorly understood \cite{hu2025neural, Zhou2025}. Recent encoding models have shown that visual, auditory, and linguistic features each explain distinct components of cortical activity during movie viewing \cite{Schrimpf2020, kamali, oota2023joint}. Yet, most approaches rely on unimodal representations or simple concatenation of modalities, leaving open whether more sophisticated, modality-specific representations and flexible fusion strategies can better model brain responses — particularly in terms of generalization beyond the training distribution.

This challenge of generalization is compounded by the fact that naturalistic stimuli like movies often feature audiovisual streams that are not perfectly aligned: dialogue, sound effects, and non-diegetic music may diverge from the visual scene. This raises critical questions about how encoding models capture variability in cross-modal structure, and whether neural responses in different brain regions are equally sensitive to such congruence.

In this work, we develop multimodal encoding models to predict parcel-level brain activity during naturalistic movie viewing, investigating the utility of combining advanced visual, auditory, and linguistic representations with alternative fusion mechanisms. We evaluate these models on a large-scale fMRI dataset, assessing prediction accuracy and generalization on both in-distribution and out-of-distribution movies that vary in audiovisual congruence.

Our contributions are as follows: 
\begin{itemize} 
	\item We show that domain-specific, pretrained feature extractors improve prediction of cortical responses, particularly in auditory regions. 
	\item We demonstrate a trade-off between model complexity and generalization by comparing linear and attention-based fusion mechanisms. 
	\item We find that prediction performance varies with audiovisual congruence of the stimulus, suggesting our models are sensitive to cross-modal structure. 
	\item We report that linguistic features did not improve predictive performance, suggesting a dominance of audiovisual streams in driving neural responses. 
\end{itemize}
Together, these findings inform the design of more robust, ecologically valid models of multimodal brain responses.
\begin{figure}[t!] 
	\centering 
	\includegraphics[width=0.8\textwidth]{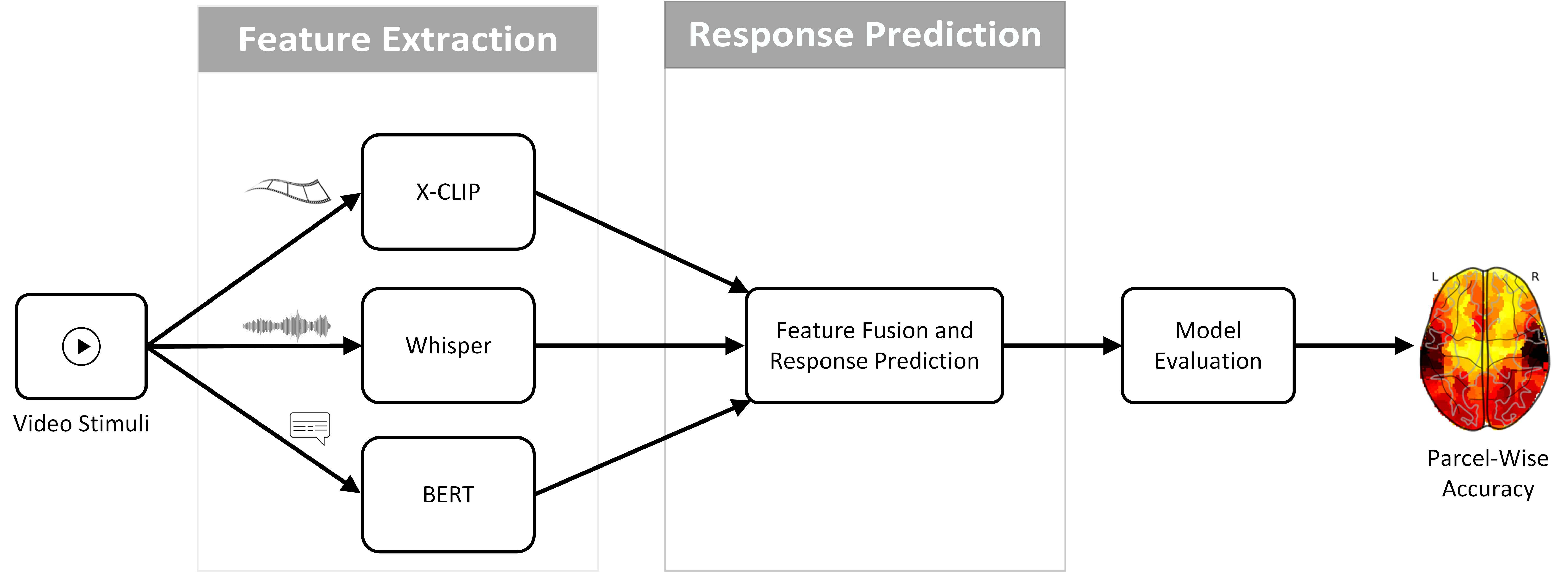} 
	\caption{A schematic of the multimodal encoding pipeline. We process naturalistic movie stimuli using state-of-the-art models to extract visual, auditory, and linguistic features. These features are then fused and mapped to predict brain activity on a parcel-wise basis. The model's predictive performance is then quantified by evaluating its predictions against actual fMRI responses.}
	\label{fig:graphical abstract} 
\end{figure}

\section{Related Works}
\label{sec:related works}
Our research is situated at the confluence of several recent advancements in modeling brain responses to naturalistic stimuli. Critical work has begun to investigate the architectural principles of multimodal fusion, with studies suggesting that models promoting early and cohesive integration of audiovisual streams may better reflect neural processing \cite{oota2025multi}. Alongside architectural choices, the quality of the underlying feature representations is paramount; indeed, it has been shown that representations aligned with neural data can improve model robustness and generalizability \cite{freteault2025alignment}. 
Other research has similarly focused on enhancing unimodal (visual) encoding models, for instance by using compression techniques \cite{kamali} or by incorporating structural priors such as the spatial relationships between neighboring voxels \cite{Ranjbar_2024}.
The practical ability to conduct such large-scale investigations is itself a recent development, enabled by computational optimizations that make training thousands of individual encoding models on massive datasets computationally feasible \cite{ahmadi2024scaling}.

Despite this progress, two critical questions often remain unaddressed. First, a model's true robustness is best revealed when it is tested on truly novel, out-of-distribution (OOD) data, a crucial step that is often omitted. Second, the relative contribution of different modalities—especially the necessity of linguistic information in the presence of rich audiovisual streams—requires more rigorous evaluation. Our study directly addresses these gaps by systematically testing different fusion models on both in-distribution and OOD data, providing a direct test of model generalization and the specific contribution of each sensory stream.

\section{Dateset and Challenge}
\label{sec:dataset and challenge}

In this work, we address the task defined by the Algonauts 2025 challenge\cite{gifford2024algonauts}, “How the Human Brain Makes Sense of Multimodal Movies”. The challenge builds on the CNeuroMod dataset \cite{boyle2023courtois}, a large-scale resource designed to capture brain responses to naturalistic, multimodal stimulation. Functional MRI data were recorded on a 3T scanner from four individuals (sub-01, sub-02, sub-03, sub-05), each contributing approximately 80 hours of whole-brain time series data with a temporal resolution (TR) of 1.49 seconds. During scanning, participants watched a diverse set of movies, presented as continuous videos with synchronized visual and auditory streams, and supplemented by time-stamped transcripts of the spoken dialogue. Moreover, the fMRI data were preprocessed and parcellated into 1,000 functional regions of interest using the Schaefer atlas \cite{schaefer2018local}.

The challenge centers on predicting these parcel-level fMRI time series from the multimodal movie inputs. Training data were drawn from the Friends sitcom (seasons $1–-6$) and a selection of movies (The Bourne Supremacy, Hidden Figures, Life, and The Wolf of Wall Street), while model performance was evaluated on two held-out test sets: Friends season 7, representing in-distribution generalization, and a separate set of previously unreleased movies to assess out-of-distribution generalization. The multimodal and ecologically valid nature of the stimuli, combined with the emphasis on out-of-distribution robustness, make this challenge a rigorous benchmark for brain encoding models.

\section{Methods}
\label{sec:methods}

This study employed a parcel-wise encoding model to map multimodal video stimuli to their corresponding fMRI BOLD responses. The video stimuli were first decomposed into three distinct modalities for analysis:
Visual (the sequence of video frames),
Auditory (the raw auditory waveform from the video's soundtrack),
and Linguistic (the textual content from time-aligned subtitles).
The encoding architecture consisted of two main stages. First, separate feature extraction modules transformed the raw data from each of the three modalities into a quantitative feature space. Second, these distinct feature sets were fused and fed into a prediction module, which was trained to learn the mapping between the combined stimulus features and the recorded BOLD activity for each brain parcel. 
The implementation of this architecture is described below. We first detail the dedicated feature extraction module for each modality, followed by the final module for feature fusion and prediction.
Notably, all analyses were implemented in Python, utilizing pretrained models accessed via the Hugging Face Hub.

\subsection{Visual Feature Extraction}
\label{subsec:visual feature extraction}

To derive feature representations from the visual stream, we used a pretrained X-CLIP model \cite{ni2022expanding}. X-CLIP is a video-language model specifically designed for temporal modeling, overcoming the limitations of static image models by integrating information across video frames. This model was selected due to its state-of-the-art performance on video understanding benchmarks, such as achieving 87.1\% accuracy on the Kinetics-400 action recognition dataset \cite{kay2017kinetics}, demonstrating its capacity to extract robust and meaningful dynamic visual features. 

For our analysis, we employed the X-CLIP-L/14 version. The movie stimuli were first segmented into non-overlapping temporal windows, where the duration of each window matched the fMRI temporal resolution (TR). From each window, 8 frames were uniformly sampled and resized to a spatial resolution of $224\times224$ pixels. These frames were then processed by the model's vision transformer backbone. Based on preliminary analyses, we selected the output of the 23rd transformer block as our visual feature space, as it offered a rich balance of semantic and temporal information.
This procedure initially yielded an $8\times257\times1024$ feature tensor for each sample. To aggregate this tensor into a single feature vector, we performed two pooling operations in sequence: first, average pooling across the 257 spatial token dimension (representing 256 image patches and 1 [CLS] token), followed by max pooling across the 8-frame dimension. This process resulted in a final 1024-dimensional feature vector for each fMRI time point.

\subsection{Auditory Feature Extraction}
\label{subsec:auditory feature extraction}

We extracted auditory representations from the audio stream using the pre-trained multilingual Whisper-base model \cite{radford2023robust}. Whisper is an encoder--decoder Transformer trained on 680{,}000 hours of weakly supervised, multilingual, and multitask audio-text pairs collected from diverse web sources. It achieves strong performance on standard automatic speech recognition (ASR) benchmarks, including 5.0\% word error rate (WER) on LibriSpeech test-clean and 12.4\% on test-other \cite{radford2023robust}, which supports its use as a high-quality speech encoder in our pipeline.

In our setup, the audio track was resampled to 16~kHz and converted to mono by averaging stereo channels. Each recording was segmented into contiguous, non-overlapping chunks of 1.49\,s, aligned with the fMRI temporal resolution. Each segment was then passed to Whisper's processor, which computes an 80-channel log-Mel spectrogram internally, and then through Whisper's convolutional stem and Transformer encoder stack. We extracted the hidden states from the final encoder layer of the encoder for each segment and averaged them over the time dimension within the segment to produce a compact feature vector aligned with each TR.

\subsection{Linguistic Feature Extraction}
\label{subsec:linguistic feature extraction}

We initially hypothesized that linguistic information from movie subtitles would improve predictions of brain activity. To test this, we extracted features using a pretrained BERT-large (uncased) model \cite{devlin2019bert}, a Transformer architecture known for generating rich, context-aware semantic representations. To capture the unfolding narrative, we employed a sliding window approach where the input for each fMRI time point was formed by concatenating text from the current and preceding samples, limited to a maximum sequence length of 512 tokens.

However, in preliminary encoding models, the inclusion of these BERT-derived features did not yield a significant improvement in predictive accuracy over a model with only visual and auditory features. Furthermore, the high dimensionality of the linguistic features substantially increased model complexity and the risk of overfitting. Therefore, to ensure a parsimonious and robust final model, the linguistic modality was excluded from all subsequent analyses reported in this paper.

\subsection{Feature Fusion and Encoding Models}
\label{subsec:Feature Fusion and Encoding Model}

Two distinct encoding models were developed to map stimulus features to parcel-wise BOLD responses: a linear model and a more complex attention-based neural network designed to capture non-linear dynamics.

Both models used an identical input preparation pipeline. To account for the delayed nature of the hemodynamic response, the input for predicting the BOLD response at a given time sample $i$ was constructed from the feature vectors of the preceding $N_{d} = 10$ time points. Based on preliminary analyses on the validation set, which indicated that linguistic features did not improve predictive accuracy and increased the risk of overfitting, these features were excluded from the final models. The subsequent analyses were therefore performed using only visual and auditory features. Prior to being fed into the models, the high-dimensional feature vectors for the visual and auditory modalities were reduced using Principal Component Analysis (PCA). The number of principal components for each modality was treated as a hyperparameter and selected to maximize prediction accuracy on the validation set.

\begin{figure}[t!] 
	\centering 
	\includegraphics[width=1\textwidth]{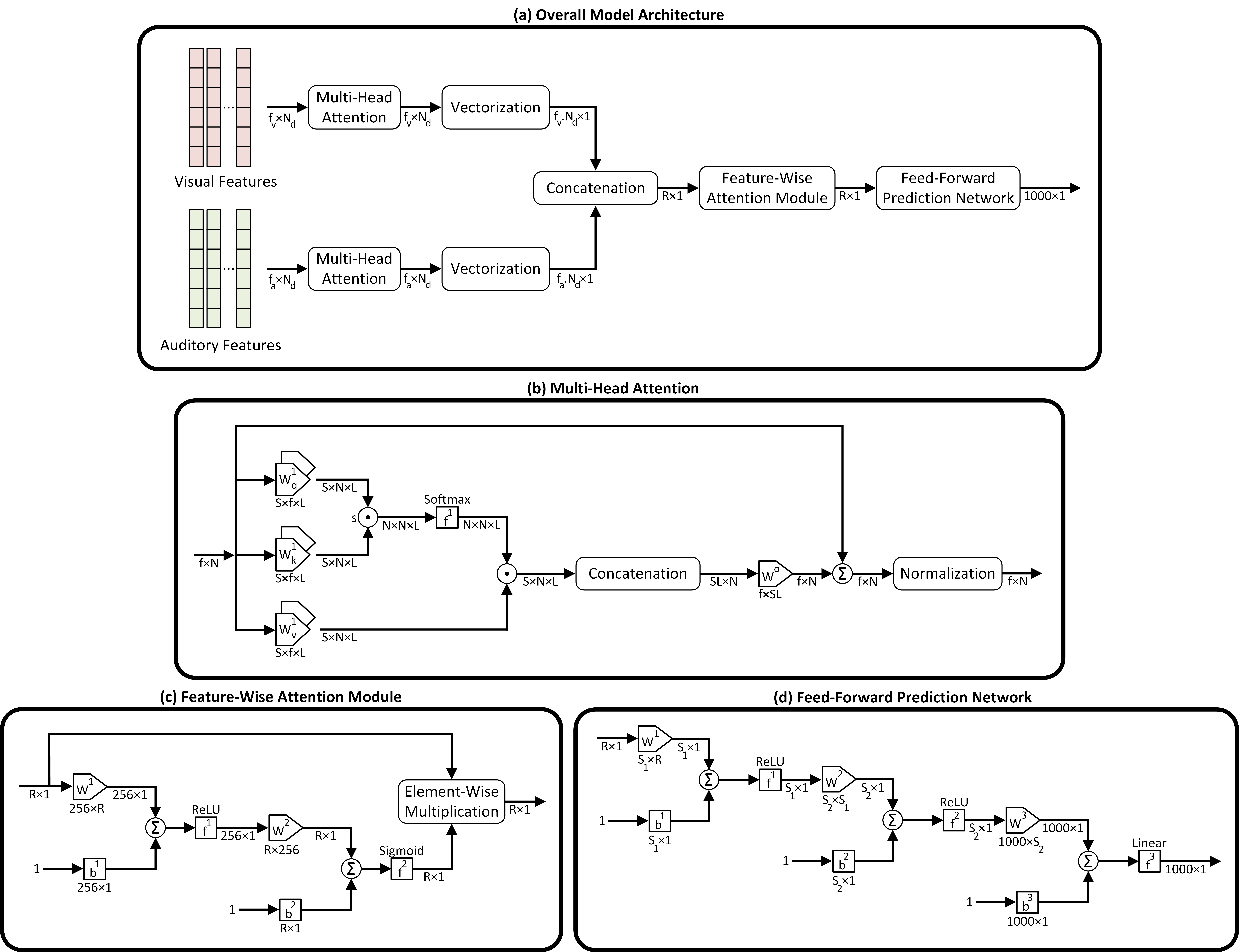} 
	\caption{The Proposed Architecture for Multimodal Brain Response Prediction. The figure details the complete architecture of the model, designed to predict fMRI activity in 1000 cortical parcels from visual and audio stimuli. 
	\textbf{(a)} Overall Model Architecture: This panel illustrates the end-to-end data pipeline. Independent visual and audio streams are first processed by modality-specific Multi-Head Attention blocks and vectorized. The streams are then fused via concatenation, refined by a Feature-wise Attention Module, and finally mapped to the target parcel activations by the Feed-Forward Prediction Network. The dimensionality of the data at each stage is indicated.
	\textbf{(b)} Multi-Head Attention Block: A detailed view of the multi-head attention mechanism used for temporal processing. The input sequence is projected into Query (Q), Key (K), and Value (V) matrices for each of the parallel attention heads. Attention weights are computed via scaled dot-product and a softmax function. The outputs are concatenated, linearly projected, and critically, combined with the original input via a residual connection before being stabilized by Layer Normalization.
	\textbf{(c)} Feature-wise Attention Module: The internal structure of the module responsible for dynamic feature fusion. It employs a two-layer MLP with a bottleneck structure and a final sigmoid activation to generate a vector of feature gates. These gates are applied to the concatenated feature vector via element-wise multiplication to dynamically recalibrate feature importance based on the input context.
	\textbf{(d)} Feed-Forward Prediction Network: The architecture of the final prediction head, implemented as a three-layer feed-forward network. It maps the fused and gated feature representation to the final 1000-parcel brain response through a series of linear transformations and ReLU activations. The final layer uses a linear activation, appropriate for the regression of BOLD signal values.}
	\label{fig:Fusion Model} 
\end{figure}

The first model, which will be called the "proposed linear model", established a linear mapping from the prepared features to the brain activity of each parcel. The input vector $p_{i}$ for a given BOLD sample $r_{i}$ was defined by concatenating the PCA-reduced feature vectors from the previous $N_{d}$ time points for both the visual ($f_{v}$) and auditory ($f_{a}$) modalities (Equation \ref{equ:input space}).
\begin{equation}
	p_{i} = concatenate\{ f_{v, i-1}, \dots, f_{v, i-N_{d}}, f_{a, i-1}, \dots, f_{a, i-N_{d}} \}
	\label{equ:input space}
\end{equation}
A separate linear regression model was trained for each of the 1000 cortical parcels. The model learns a weight vector $w$ and a bias $b$ to predict the response in Equation \ref{equ:linear regression}.
\begin{equation}
	r_{i} = w^T p_{i} + b
	\label{equ:linear regression}
\end{equation}
The cost function to estimate the weight vector from training dataset is presented in Equation \ref{equ:linear regression - cost function}. The model weights were estimated by minimizing the sum of squared errors. 
\begin{equation}
	f(w) = || r_{i} - w.p_{i} ||^2_2
	\label{equ:linear regression - cost function}
\end{equation}
To model more complex temporal and cross-modal interactions, we developed a neural network architecture, depicted in Figure \ref{fig:Fusion Model}. The network processes temporal dynamics within each modality before fusing them for a final prediction and is composed of three primary stages. This model will be called the "proposed attention-based model".

First, the input sequences for the visual and auditory modalities were processed by separate Multi-Head Self-Attention blocks. These blocks identified complex temporal patterns by dynamically weighing the importance of different time steps within the $N_{d}$ input window for each modality. Each block included residual connections and layer normalization to ensure stable training. 

Second, the processed modality representations were concatenated and passed to a Feature-wise Attention Module. This module, analogous to a Squeeze-and-Excitation block \cite{hu2018squeeze}, generated an input-specific gating vector via a two-layer MLP with a bottleneck structure and a sigmoid activation. This gate was applied via element-wise multiplication to the concatenated features, allowing the model to dynamically recalibrate the contribution of each feature from both modalities. 

Finally, the fused and recalibrated feature vector was mapped to the 1000-dimensional brain response space using a Feed-Forward Prediction Network. This network consisted of a three-layer MLP with ReLU activations and dropout layers for regularization. The final linear output layer produced the predicted BOLD signal for each of the 1000 cortical parcels.

\section{Results and Discussion}
\label{sec:results and discussion}

This section evaluates the performance of our proposed multimodal encoding models and discusses the implications of our findings for understanding how the brain processes and integrates information from different modalities.

\subsection{Evaluation Metric}
\label{subsec:Evaluation Metric}

To quantify the predictive accuracy of our encoding models, we employed the Pearson correlation coefficient. This metric measures the linear correlation between the predicted and actual blood-oxygen-level-dependent (BOLD) fMRI signals for each brain parcel. The formula is presented in Equation \ref{equ:evaluation metric}, where $\rho_{i}$ is the Pearson correlation for parcel $i$, $N_s$ is the number of temporal samples in the BOLD signal, $r_{i}$ represents the actual BOLD response, and $r_{i, pred}$ is the predicted response. The terms $\bar{r}_{i}$ and $\bar{r}_{i, pred}$ denote the mean of the actual and predicted responses, respectively.
\begin{equation}
	\rho_{i} = \frac{\sum _{n=1}^{N_s} (r_{i, pred}(n) - \bar{r}_{i, pred})(r_{i}(n) - \bar{r}_{i})}{\sqrt{ \sum _{n=1}^{N_s} (r_{i, pred}(n) - \bar{r}_{i, pred})^2 } \sqrt{\sum _{n=1}^{N_s} (r_{i}(n) - \bar{r}_{i})^2}}
	\label{equ:evaluation metric}
\end{equation}

\subsection{Model Performance on In-Distribution Data}
\label{subsec:in-distribution accuracy}

We first assessed our models' performance on an in-distribution (ID) test set, which consisted of the seventh season of the Friends television series. This dataset shares critical low-level and high-level properties with the training corpus (Seasons 1-6), including the same characters, narrative structure, and overall cinematic style.

The primary purpose of this ID evaluation is to confirm that the model has learned a robust mapping between the multimodal features and the corresponding fMRI BOLD responses, rather than simply memorizing the training examples. Strong predictive performance on this unseen data from a familiar domain is a prerequisite for a successful model. It validates the model's ability to generalize over the temporal continuity of this specific environment, serving as a fundamental test of its learning efficacy before proceeding to more stringent evaluations.

As fully discussed in Section \ref{sec:methods}, our proposed architecture comprises specialized feature extractors for each modality: X-Clip for visual information, Whisper for auditory signals, and BERT for linguistic content. These features were then integrated and mapped to brain activity using either a linear regression model or a more complex attention-based fusion mechanism. We compared these two models against a baseline model that utilized SlowFast-R50 \cite{feichtenhofer2019slowfast}, Mel-Frequency Cepstral Coefficients (MFCCs), and a base BERT model \cite{devlin2019bert} for feature extraction, followed by a simple concatenation and linear regression \cite{gifford2024algonauts}.

\begin{figure}[t!] 
 	\centering 
 	\includegraphics[width=0.8\textwidth]{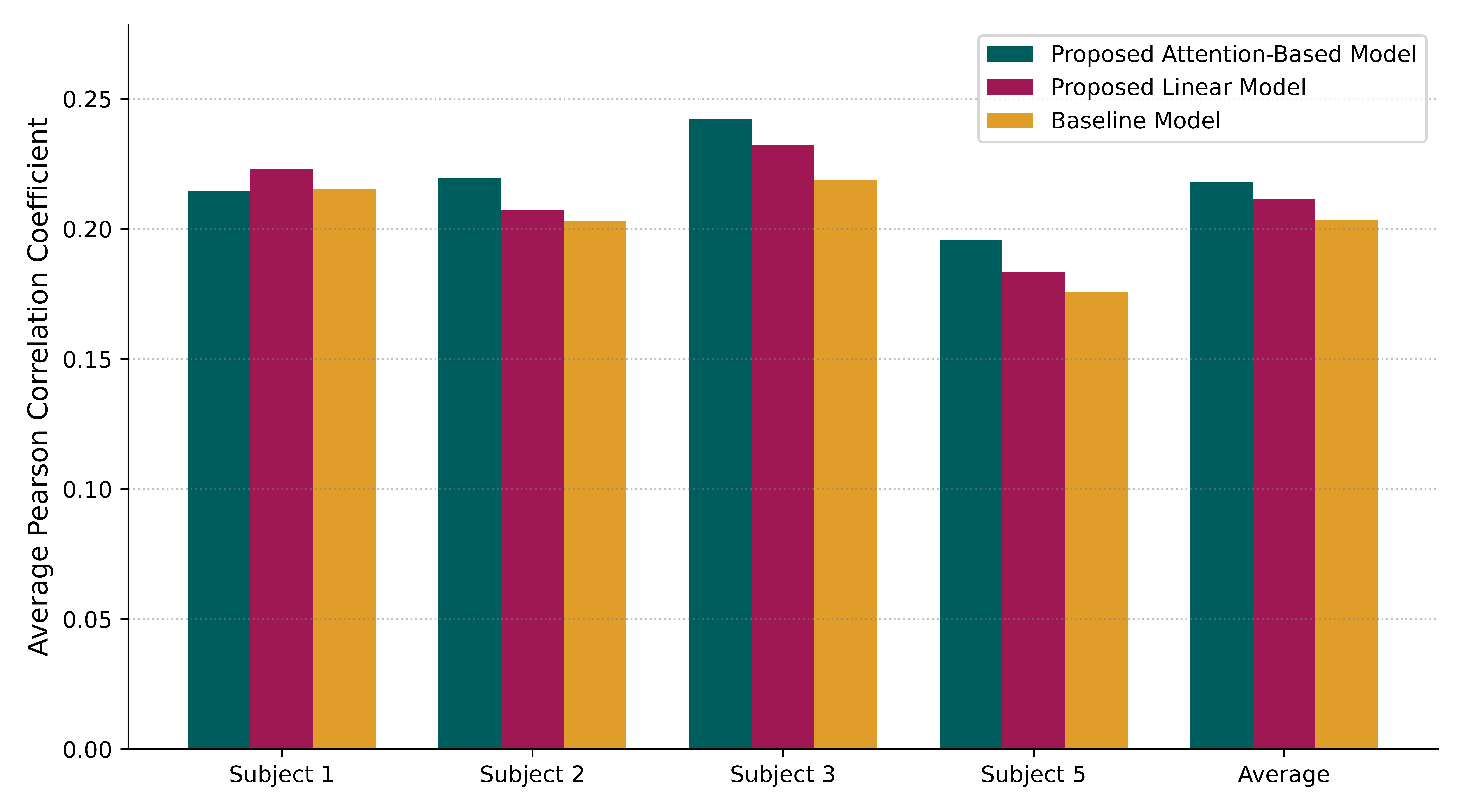} 
 	\caption{Comparative analysis of multimodal encoding model accuracy on the in-distribution test dataset. The bar graph displays the average Pearson correlation coefficient across all brain parcels for each subject, comparing our two proposed models with the baselin}
 	\label{fig:accuracy - in-distribution test} 
 \end{figure}

Figure \ref{fig:accuracy - in-distribution test} presents a comparison of the average Pearson correlation across 1000 brain parcels for each of the four subjects. The attention-based model achieved the highest average correlation (0.218), outperforming the proposed linear model (0.212) and the baseline (0.203). This represents a 7\% and 4\% improvement over the baseline for the attention-based and linear models, respectively. Notably, the attention-based model demonstrated superior performance for three out of the four subjects.

\begin{figure}[t!] 
	\centering 
	\includegraphics[width=0.8\textwidth]{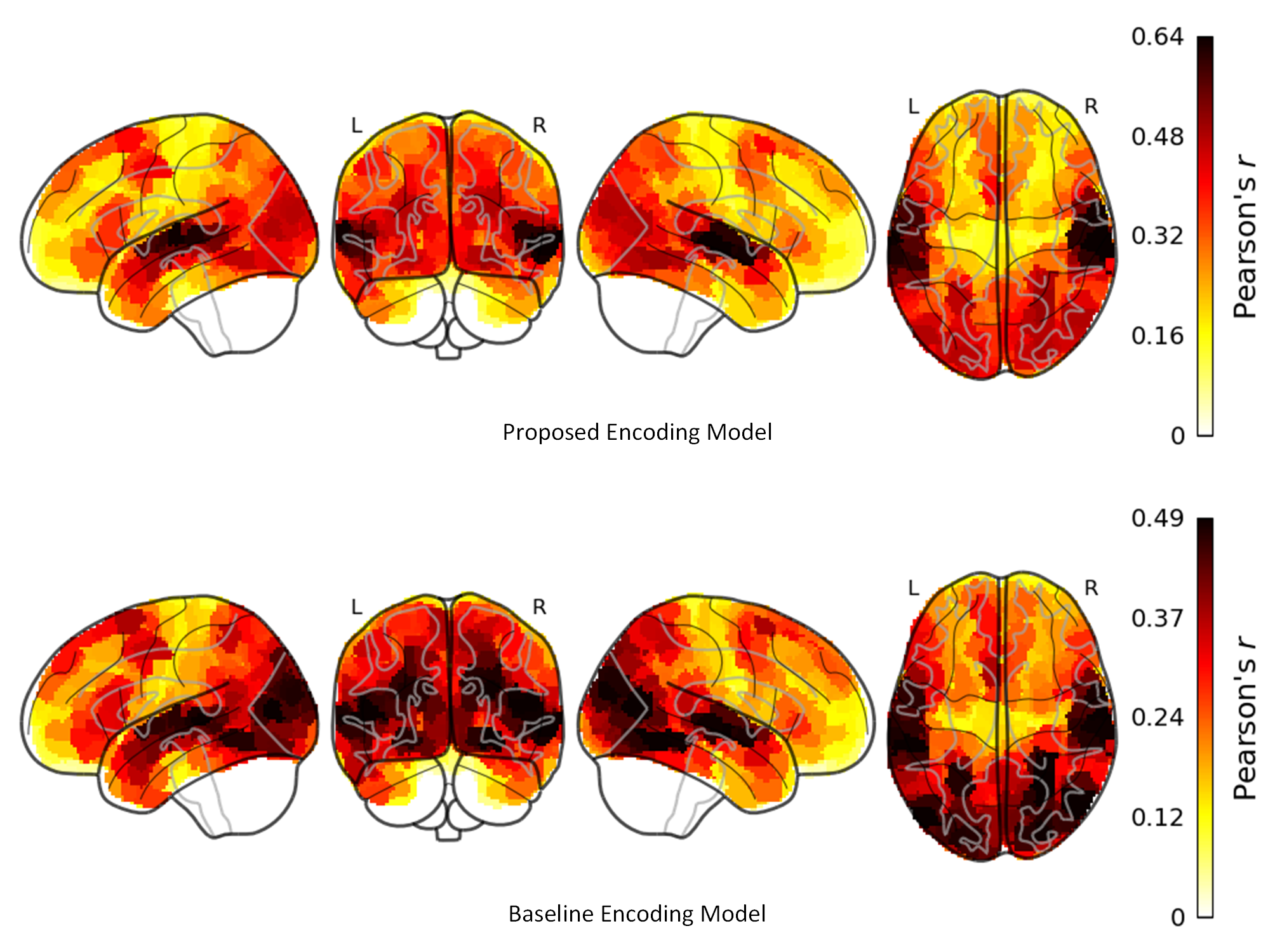} 
	\caption{Glass brain visualization of the average Pearson correlation coefficient across subjects for the in-distribution test dataset. The top plot illustrates the performance of the attention-based encoding model, while the bottom plot shows the accuracy of the baseline model. Note that the color bars indicate different correlation ranges.}
	\label{fig:spatial mapping - in-distribution test} 
\end{figure}

A spatial analysis of the model predictions, visualized as glass brain projections in Figure \ref{fig:spatial mapping - in-distribution test}, reveals distinct patterns of performance. The top panel shows the average correlation for our attention-based model across subjects, while the bottom panel displays the results for the baseline. Our model shows a marked improvement in the auditory cortex, which we attribute to the advanced speech representation capabilities of the Whisper model. In contrast, the performance in the visual cortex was comparable to the baseline.

\subsection{Generalization to Out-of-Distribution Data}
\label{subsec:out-of-distribution accuracy}

To rigorously test the models' ability to generalize to truly novel contexts, we evaluated them on an out-of-distribution (OOD) test set. This dataset comprised a diverse collection of six movies and documentaries, featuring significant stylistic and content-based shifts from the training data, including variations in cinematography, genre, and narrative structure. Success on this OOD set is a critical test of whether our models have learned fundamental principles of neural encoding rather than merely memorizing the characteristics of the training stimulus.

\begin{figure}[h!] 
	\centering 
	\includegraphics[width=0.8\textwidth]{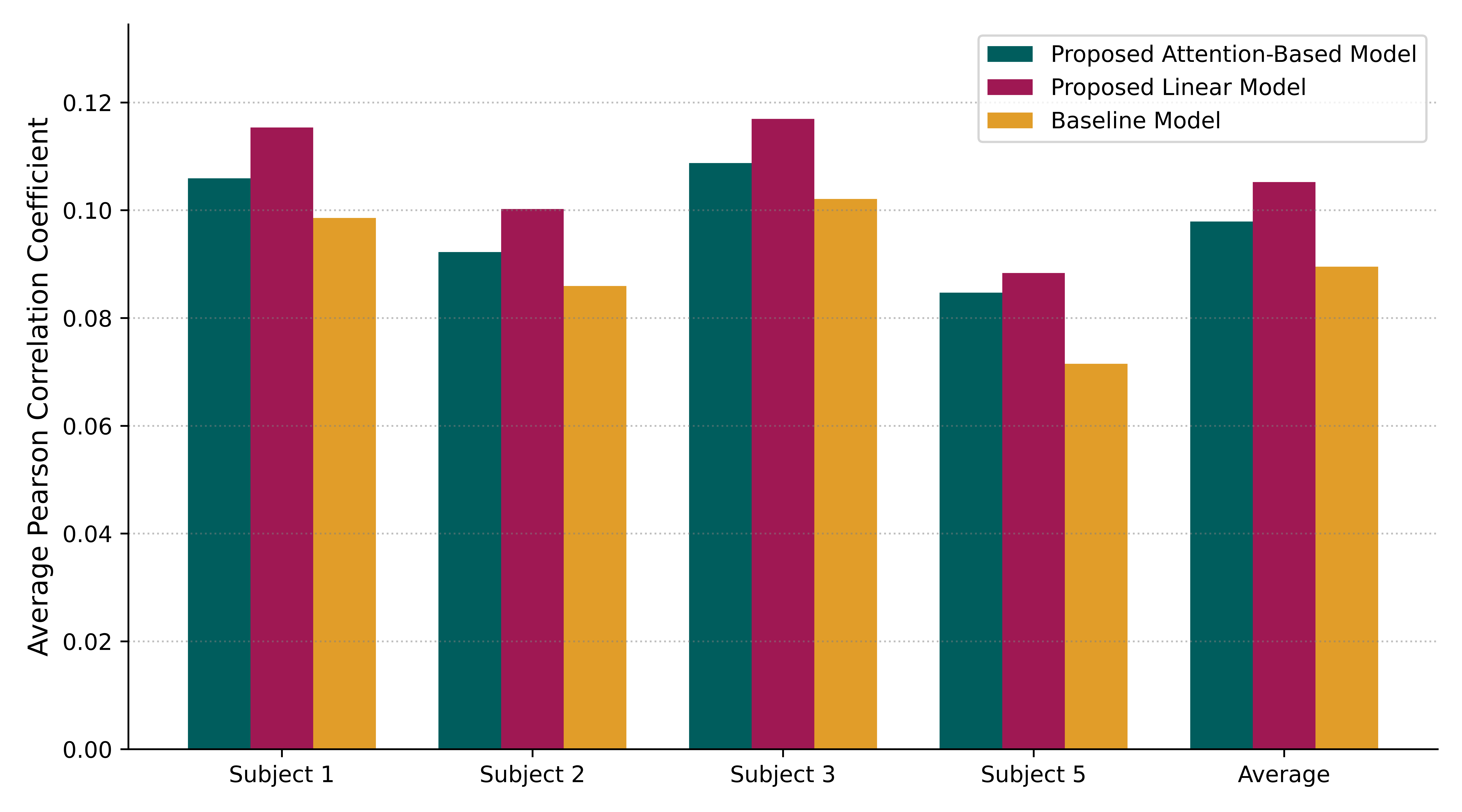} 
	\caption{Comparative analysis of multimodal encoding model accuracy on the out-of-distribution test dataset. The bar graph presents the average Pearson correlation coefficient across all brain parcels, comparing our proposed models with the baseline.}
	\label{fig:accuracy - out-of-distribution test} 
\end{figure}

The results on the OOD dataset, presented in Figure \ref{fig:accuracy - out-of-distribution test}, reveal a different pattern of performance. Here, the simpler linear model demonstrated the best generalization, achieving an average Pearson correlation of 0.105, an 18\% improvement over the baseline. The attention-based model, with an average correlation of 0.098, still outperformed the baseline by 9\%. The linear model's superiority was consistent across all four subjects in the OOD setting.

This reversal in performance between the ID and OOD tests suggests a trade-off between model complexity and generalization. While the attention-based model's higher capacity allowed it to better fit the nuances of the Friends dataset, it appears to have overfit to the specific stylistic and narrative patterns of the training data. The simpler linear model, in contrast, learned a more robust and generalizable mapping from stimulus features to brain activity.

\begin{figure}[h!] 
	\centering 
	\includegraphics[width=0.8\textwidth]{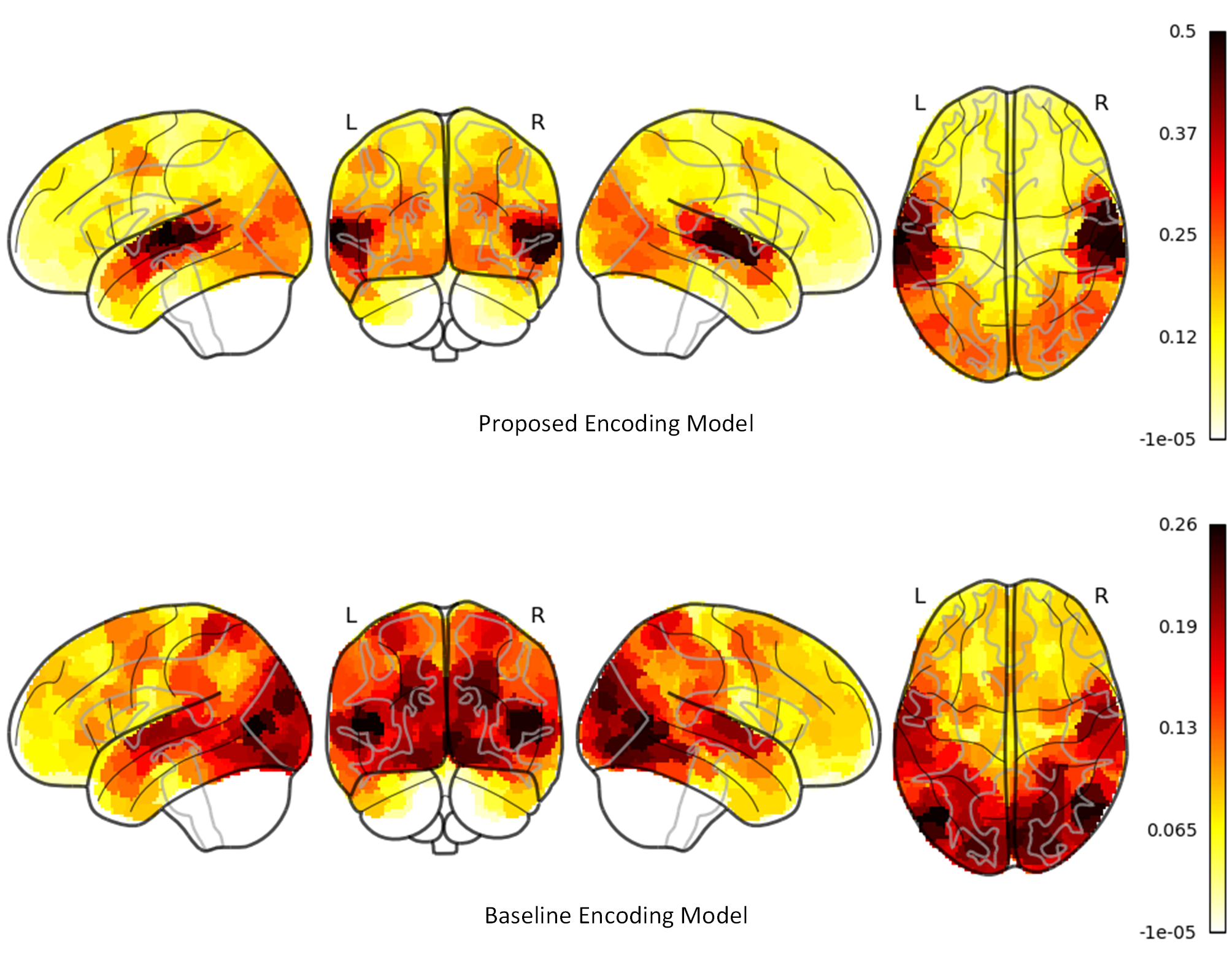} 
	\caption{Glass brain visualization of the average Pearson correlation coefficient across subjects for the out-of-distribution test dataset. The top plot illustrates the performance of the proposed linear encoding model, while the bottom plot depicts the accuracy of the baseline model. Note the different ranges on the color bars.}
	\label{fig:spatial mapping - out-of-distribution test} 
\end{figure}

The spatial accuracy on the OOD data, shown in Figure \ref{fig:spatial mapping - out-of-distribution test}, further supports this interpretation. Our linear model maintained a competitive performance in the visual cortex and continued to show significant advantages in the auditory cortex compared to the baseline, highlighting the consistent benefit of the Whisper-based audio features.

An analysis of the linear model's performance across the individual OOD stimuli (Table \ref{tab:accuracy - out-of-distribution test}) provides further insights. The model performed worst on a silent, black-and-white Chaplin film, which is unsurprising given the profound difference in visual and auditory features compared to the training data. Interestingly, the model achieved its highest performance in the auditory cortex when processing the Planet Earth documentary, a stimulus characterized by a strong correspondence between the narrative audio and the visual content. This suggests that our model is particularly sensitive to the congruent integration of auditory and visual information.

\begin{table}[h] 
	\caption{Encoding model performance on the out-of-distribution test set. The table displays the average Pearson correlation coefficient across all subjects for each of the six movies, as predicted by the proposed linear model.}
	\centering 
	\begin{tabular}{lc}
		\toprule
		\textbf{OOD Stimulus (Movie)} & \textbf{Average Pearson Correlation Coefficient} \\
		\midrule
		Chaplin          & 0.058 \\
		Princess Mononoke     & 0.130  \\
		Le Passe-Partout             & 0.118 \\
		Planet Earth                 & 0.094 \\
		Pulp Fiction                 & 0.137 \\
		The Wheel of Time            & 0.094 \\
		\midrule
		\textbf{Overall Average}     & \textbf{0.105} \\
		\bottomrule
	\end{tabular}
	\label{tab:accuracy - out-of-distribution test}
\end{table}

The results of this study offer several key insights into the neural encoding of multimodal stimuli. First, the superior performance of our models, particularly in the auditory cortex, underscores the importance of using powerful, domain-specific feature extractors. The success of the Whisper model suggests that representations from models pre-trained on large-scale, diverse audio data are highly effective at capturing the neural responses to complex auditory scenes.

Second, the exclusion of linguistic features from our final models—prompted by their lack of predictive improvement and increased overfitting risk—is itself an intriguing finding. We speculate that this may reflect the hierarchical nature of sensory processing during naturalistic perception. For native speakers comprehending a familiar language, the rich, continuous information from the visual and auditory streams may be the dominant drivers of neural activity, rendering the discrete, symbolic information from subtitles largely redundant. While this result could also stem from model-specific limitations, it raises the compelling hypothesis that the brain dynamically prioritizes information streams based on their ecological salience and informational necessity.

Third, the contrasting performance of the linear and attention-based models on the ID and OOD datasets highlights a critical trade-off in neural encoding models. While more complex, higher-capacity models may achieve better performance on data that is similar to the training set, they are also more prone to overfitting and may not generalize as well to novel stimuli. This finding has important implications for the design of future encoding models, suggesting that a balance must be struck between model complexity and the learning of robust, generalizable representations.

\begin{figure}[t!] 
	\centering 
	\includegraphics[width=1\textwidth]{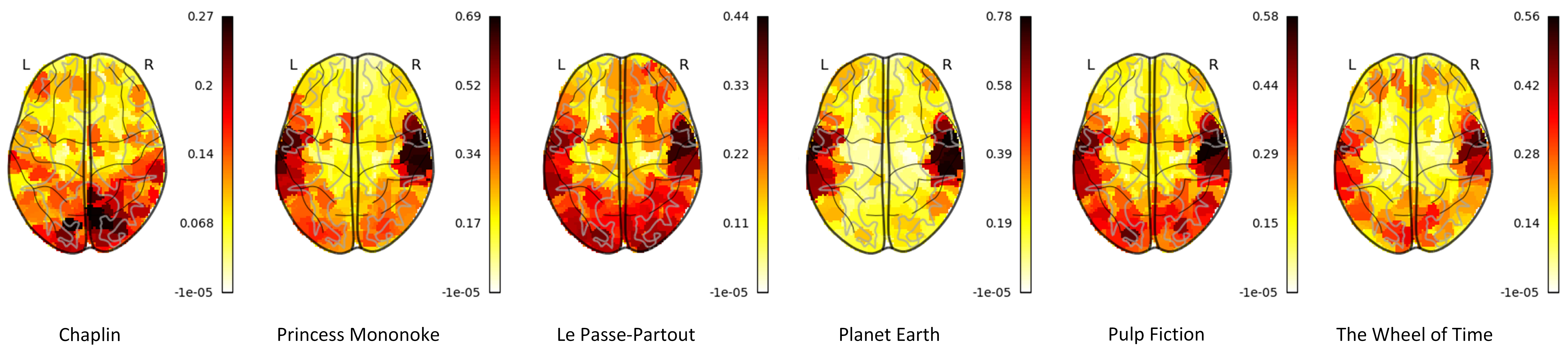} 
	\caption{Spatial distribution of model performance across the six out-of-distribution (OOD) stimuli for the second subject. Each glass brain projection visualizes the Pearson correlation coefficient between predicted and actual fMRI signals for a specific movie. Results are shown for the proposed linear model. The figure highlights the stimulus-dependent nature of generalization, with predictive accuracy varying considerably across the different cinematic contexts.}
	\label{fig:ood performance sub2} 
\end{figure}

Fourth, the variable performance across different OOD stimuli provides clues about the nature of multimodal integration in the brain. The strong performance in auditory cortex on the Planet Earth documentary, where the audio directly describes the visual content, suggests that the brain regions captured by our model are highly responsive to congruent audiovisual information. Conversely, the poor performance in auditory cortex on the Chaplin film indicates that the model is less adept at capturing neural responses to stimuli that lack the familiar audio-visual statistical regularities of the training data (Look at Figure \ref{fig:ood performance sub2}).

In conclusion, the dual-testing approach on both in-distribution and out-of-distribution data provides a more nuanced and rigorous evaluation of our models' capabilities. The findings not only demonstrate the effectiveness of our proposed architecture but also shed light on the complex interplay between model design, stimulus characteristics, and the neural mechanisms of multimodal perception. Future work should explore more sophisticated fusion mechanisms that can capture the dynamic and context-dependent nature of multimodal integration in the brain, while also being robust to shifts in the input distribution.


\bibliographystyle{unsrt}  
\bibliography{references}

\end{document}